\documentclass{article}

    \usepackage[preprint, nonatbib]{neurips_2021}
    
\usepackage[utf8]{inputenc} 
\usepackage[T1]{fontenc}    
\usepackage{hyperref}       
\usepackage{url}            
\usepackage{booktabs}       
\usepackage{amsfonts}       
\usepackage{nicefrac}       
\usepackage{microtype}      
\usepackage{xcolor}         
\usepackage{amsmath}
\usepackage{tikz}
\usepackage{mathdots}
\usepackage{yhmath}
\usepackage{cancel}
\usepackage{color}
\usepackage{siunitx}
\usepackage{array}
\usepackage{multirow}
\usepackage{amssymb}
\usepackage{tabularx}
\usepackage{extarrows}
\usepackage{booktabs}
\usetikzlibrary{fadings}
\usetikzlibrary{patterns}
\usetikzlibrary{shadows.blur}
\usetikzlibrary{shapes}
\usepackage{bm}

\title{SubOmiEmbed: Self-supervised Representation Learning of Multi-omics Data for Cancer Type Classification}

\author{%
  Sayed Hashim \qquad Muhammad Ali \qquad Karthik Nandakumar \qquad Mohammad Yaqub\\
  Mohamed Bin Zayed University of Artificial Intelligence, UAE\\
  \texttt{\{sayed.hashim, muhammad.ali, karthik.nandakumar, mohammad.yaqub\}@mbzuai.ac.ae} \\
}

\begin{document}
\nocite{*}
\maketitle

\begin{abstract}
For personalized medicines, very crucial intrinsic information is present in high dimensional omics data which is difficult to capture due to the large number of molecular
features and small number of available samples. Different types of omics data show various aspects of samples. Integration and analysis of multi-omics data give us a broad view of tumours, which can improve clinical decision making. 
Omics data, mainly DNA methylation and gene expression profiles are usually high dimensional data with a lot of molecular features. 
In recent years, variational autoencoders (VAE) \cite{vae} have been extensively used in embedding image and text data into lower dimensional latent spaces. 
In our project, we extend the idea of using a VAE model for low dimensional latent space extraction with the self-supervised learning technique of feature subsetting. With VAEs, the key idea is to make the model learn meaningful representations from different types of omics data, which could then be used for downstream tasks such as cancer type classification. The main goals are to overcome the curse of dimensionality and integrate methylation and expression data to combine information about different aspects of same tissue samples, and hopefully extract biologically relevant features. Our extension involves training encoder and decoder to reconstruct the data from just a subset of it. By doing this, we force the model to encode most important information in the latent representation. We also added an identity to the subsets so that the model knows which subset is being fed into it during training and testing. We experimented with our approach and found that SubOmiEmbed produces comparable results to the baseline OmiEmbed \cite{omiembed} with a much smaller network and by using just a subset of the data. This work can be improved to integrate mutation-based genomic data as well.
 \\
\end{abstract}

\section{Introduction}
The word omics refers to areas of studies in biological sciences which end with omics, such as genomics, proteomics, transcriptomics, or metabolomics \cite{omics}. The advent of Next Generation Sequencing (NGS) has brought about huge amounts of multi-omics data. Different types of omics
data show various aspects of samples. Omics data, mainly DNA methylation and gene expression profiles are usually high dimensional data with a lot of molecular features from less samples.  This makes it hard to extract insights from them. This is called 'the curse of dimensionality' in machine learning.

Omics profiling usually tells us the differences associated with a disease from an omics point of view. Such profiling can be used as markers of the disease, in extracting insights about the biological pathways of the disease as well as about treatment options \cite{multiomics}. Moreover, integration and analysis of multi-omics data give us a broader view of tumours, which can improve clinical decision making, as each type of profiling reveals particular characteristics of the tumours. Therefore, it is safe to say that extracting information through intelligent computational tools can go a long way in improving patient outcomes.

Huge number of molecular features of less number of samples makes it challenging to build computational tools to extract insights from genome-wide high dimensional multi-omics data. As most of the omics profiling nowadays is done genome-wide, the dimensionality of such data is pretty large. For instance, RNA-Seq gene expression profiling comprises of more than 60,000 values for a single sample, while DNA methylation profiling contains more than 485,000 values for a sample. At the same time, the number of samples available in a multi-omics dataset is usually less, because of the time and resources involved in collecting samples from patients. Curse of dimensionality, can cause huge overfitting of a model and make samples harder to group. 

In this project, we use a deep embedding network that learns to represent multi-omics data types in a latent space with lower dimension. We implemented the baseline called OmiEmbed \cite{omiembed} by Zhang et al. that uses a variational autoencoder for low dimensional latent space extraction by integrating genome-wide miRNA \& gene expression as well as DNA methylation profiles. The key idea here is to make the model learn meaningful representations from different types of omics data, which could then be used for downstream tasks such as cancer type classification. The main goals here are to overcome the curse of dimensionality in multi-omics data, combine information about different aspects of samples, and hopefully extract biologically relevant features.

We also explore the field of Self-Supervised Learning (SSL) in this project. SSL has been successfully used in recent years to learn meaningful representations of data in natural language processing \cite{nlp1, nlp2} and image domains \cite{image1, image2}. It is usually done by taking advantage of temporal, semantic, spatial or structural relationships in the data. But such methods are less effective due to the absence of such relationships in tabular data. Augmentation methods like color transformation, affine and projective transformations etc. are domain
specific, and not suited to tabular setting. Due to the above reasons, the domain of SSL has not been explored much in tabular setting compared to other domains.

In this project, we divide the features into subsets and train the model to reconstruct all features from a single subset of features, thus making the model learn from multiple views of the dataset. This is similar to feature bagging in ensemble learning or cropping in image domain. Training the model to reconstruct the data from a subset of its features results in the encoder learning better representations. During testing, representations from multiple subsets are aggregated. Additionally, we explore adding an identity for the subset while training for downstream tasks. This helps the model identify the subset of features that is fed into it. Thus the model learns the corresponding view during training and uses this information during testing.

\section{Literature review}
The success of representation learning ability of DNNs (deep neural networks) in computer vision and natural language processing motivated researchers to apply DNNs  in various downstream tasks using  high-dimensional omics data. These tasks include  classification of diseases, predicting age and others.
For classification and discrimination of breast tumour samples from benign samples, Danaee et al. \cite{Danaee2017ADL} proposed  a new model used for  cancer detection where they utilized gene expression data  by using autoencoder with stacked denoising. In another work, Lyu and Haque \cite{Lyu} used convolutional neural network for tumour type classification. For this purpose, initially images were obtained from  RNA-Seq high dimensional data which achieved an accuracy of 95.59\%. 

Graph convolution neural network (GCN) along with relation network (RN) were proposed by Rhee et al. \cite{rhee} to classify subtypes of breast tumours. In this method, gene expression data along with  protein-protein interaction networks were used. Zhang et al. proposed OmiVAE \cite{omivae} (variational autoencoder based generative model strategy) for tumour type classification and it achieved an accuracy of 97.49\%  from  33 tumour categories and normal class using gene expression as well as DNA methylation data from GDC dataset. 

Apart from classification tasks, scientist also tried to use deep neural networks for various other downstream tasks using high dimensional multi-omics data. For the prediction of survival for 20 different tumour types a multi-modal approach was proposed by Cheerla and Gevaert \cite{cheerla} using  GDC dataset.  Along with multi-omics data, they  used additional clinical information \& histopathology whole slide images (WSIs) and achieved an overall C-index of 0.78. To further explore in this direction, an innovative deep neural framework called  SAUCIE was proposed by Amodio et al. \cite{saucie}. In this work, they used single cell gene expression data to perform imputation, clustering, batch correction, denoising, and visualisation. The state-of-the-art in this domain is OmiEmbed \cite{omiembed} (to the best of our understanding), which we implemented as baseline.

Corrupting the data by adding noise is a common SSL approach in tabular data \cite{denauto}. In this approach, an autoencoder is trained to represent corrupted examples of data on a latent space, after which it is trained to reconstruct the uncorrupted data from this latent representation. Thus, the model is trained to learn representations that are robust to the noise in the input This type of autoencoder is called a denosing autoencoder. A drawback of this approach is that it makes the assumption that all features are equally important and treats them equally. But, adding noise to uninformative features may not help in this method. 

Another work called VIME (Value Imputation and Mask Estimation) \cite{VIME} uses SSL on tabular data by training a denoising autoencoder on an additional pretext task of predicting mask vectors from corrupted data on top of reconstructing the data. To predict which features have been masked in mask estimation task, a classifier connected to the latent representation layer is trained on corrupted data. But, this approach still relies on noisy data in the input as in the case of a standard denoising autoencoder. Moreover, training a classifier on an imbalanced binary mask for large-dimensional data might not the make the model learn meaningful representations.

A recent approach named SubTab \cite{subtab} (Subsetting features of Tabular data) reduces the reliance on noise by changing the task from learning from tabular data to a multi-view representation learning problem through
subsetting the features, i.e., dividing the input features into subsets. They show that reconstructing the input data from the subset of its features instead of corrupted data in can make the autoencoder learn better latent representations. At test time, the joint representation is calculated as the aggregate of latent representations of each subset, which they call collaborative inference.

An approach named TabNet \cite{tabnet} uses sequential attention to decide the features to use at each decision step. It performs unsupervised representation learning through masked SSL. This provides a better encoder model for the supervised learning task. In a recent work on continual learning on multiple tasks \cite{continual}, authors show that task identity can be used to avoid catastrophic forgetting when the model is trained sequentially on multiple tasks. We take inspiration from this work and add subset identity in our framework. 
\section{Methods and Materials}
\subsection{Dataset}
For our experiments, we used The Cancer Genome Atlas (TCGA) pan-cancer multi-omics dataset \cite{17}. An overview of the dataset is given in Table \ref{tab:tcga}. It is one of the most popular multi omics dataset. It consists of omics data as well as phenotypic information of patients. We used 3 different omics data from the TCGA dataset, namely, DNA methylation, miRNA stem loop expression as well as gene expression. They are 485,577, 1881 and 60,483 dimensional respectively. The dataset contains samples of 33 different tumour types, as well as of normal tissues.

\begin{table}[h]
\label{tab:tcga}
\begin{tabular}{@{}lccc@{}}
\toprule
\textbf{Dataset} & \multicolumn{3}{c}{\textbf{TCGA}}                    \\ \midrule
Domain           & \multicolumn{3}{c}{Pan-cancer}                       \\
Tumour types     & \multicolumn{3}{c}{33 + 1(normal) = 34}              \\
Omics data type  & Gene expression & DNA methylation & miRNA expression \\
No of features   & 60,483          & 485,577         & 1881             \\
No of samples    & 11,538          & 9736            & 11,020           \\ \bottomrule
\\
\end{tabular}
\centering
\caption{An overview of TCGA pan-cancer dataset}
\end{table}

\subsection{Data preprocessing}
We downloaded harmonised data of 3 different omics from \href{https://xenabrowser.net/datapages/}{UCSC Xena data portal} \cite{ucsc}. RNA-Seq gene expression dataset contains 60,483 identifiers, one for each gene. Gene expression level is quantified as the $\log_2$ transformation of fragments per kilobase of transcript per million mapped reads (FPKM) value. In the case of miRNA expression dataset, the expression levels of miRNA stem-loop features were quantified as the $\log_2$ transformation of reads per million mapped reads (RPM) value. 

DNA methylation dataset consists of beta values for each CpG site and represents the ratio of methylated to total array intensity for the corresponding CpG site \cite{gdc}. Lower values of beta imply lower levels of methylation and vice-versa. The missing beta values in DNA methylation dataset were imputed using the mean of beta values of the respective CpG sites. All three datasets were rescaled so that the vector of each sample has unit norm, independent of the sample's distribution.
\subsection{Architecture}

\tikzset{every picture/.style={line width=0.75pt}} 

\begin{figure}
\centering
\begin{tikzpicture}[x=0.75pt,y=0.75pt,yscale=-1,xscale=1]

\draw  [fill={rgb, 255:red, 74; green, 144; blue, 226 }  ,fill opacity=1 ] (158.86,50.08) -- (198.91,61.92) -- (199.09,107.92) -- (159.14,120.08) -- cycle ;
\draw  [fill={rgb, 255:red, 74; green, 144; blue, 226 }  ,fill opacity=1 ] (158.86,137.08) -- (198.91,148.92) -- (199.09,194.92) -- (159.14,207.08) -- cycle ;
\draw  [fill={rgb, 255:red, 74; green, 144; blue, 226 }  ,fill opacity=1 ] (158.86,227.08) -- (198.91,238.92) -- (199.09,284.92) -- (159.14,297.08) -- cycle ;
\draw  [fill={rgb, 255:red, 248; green, 231; blue, 28 }  ,fill opacity=1 ] (257,122) -- (277,122) -- (277,219) -- (257,219) -- cycle ;
\draw    (200,87) -- (228.54,129.31) -- (256.88,171.34) ;
\draw [shift={(258,173)}, rotate = 236] [color={rgb, 255:red, 0; green, 0; blue, 0 }  ][line width=0.75]    (10.93,-3.29) .. controls (6.95,-1.4) and (3.31,-0.3) .. (0,0) .. controls (3.31,0.3) and (6.95,1.4) .. (10.93,3.29)   ;
\draw    (200,173) -- (222,173) -- (256,173) ;
\draw [shift={(258,173)}, rotate = 180] [color={rgb, 255:red, 0; green, 0; blue, 0 }  ][line width=0.75]    (10.93,-3.29) .. controls (6.95,-1.4) and (3.31,-0.3) .. (0,0) .. controls (3.31,0.3) and (6.95,1.4) .. (10.93,3.29)   ;
\draw    (199,264) -- (229,218) -- (256.92,174.68) ;
\draw [shift={(258,173)}, rotate = 122.8] [color={rgb, 255:red, 0; green, 0; blue, 0 }  ][line width=0.75]    (10.93,-3.29) .. controls (6.95,-1.4) and (3.31,-0.3) .. (0,0) .. controls (3.31,0.3) and (6.95,1.4) .. (10.93,3.29)   ;
\draw  [fill={rgb, 255:red, 245; green, 166; blue, 35 }  ,fill opacity=1 ] (305,107) -- (320,107) -- (320,159) -- (305,159) -- cycle ;
\draw  [fill={rgb, 255:red, 245; green, 166; blue, 35 }  ,fill opacity=1 ] (305,183) -- (320,183) -- (320,235) -- (305,235) -- cycle ;
\draw    (276,176) -- (291,152) -- (303.81,134.61) ;
\draw [shift={(305,133)}, rotate = 126.38] [color={rgb, 255:red, 0; green, 0; blue, 0 }  ][line width=0.75]    (10.93,-3.29) .. controls (6.95,-1.4) and (3.31,-0.3) .. (0,0) .. controls (3.31,0.3) and (6.95,1.4) .. (10.93,3.29)   ;
\draw    (276,176) -- (291,196) -- (301.85,211.37) ;
\draw [shift={(303,213)}, rotate = 234.78] [color={rgb, 255:red, 0; green, 0; blue, 0 }  ][line width=0.75]    (10.93,-3.29) .. controls (6.95,-1.4) and (3.31,-0.3) .. (0,0) .. controls (3.31,0.3) and (6.95,1.4) .. (10.93,3.29)   ;
\draw  [fill={rgb, 255:red, 233; green, 120; blue, 134 }  ,fill opacity=1 ] (343,141.33) -- (358,141.33) -- (358,193.33) -- (343,193.33) -- cycle ;
\draw  [fill={rgb, 255:red, 248; green, 231; blue, 28 }  ,fill opacity=1 ] (386,121.67) -- (406,121.67) -- (406,218.67) -- (386,218.67) -- cycle ;
\draw    (357.67,169.67) -- (383.67,169.67) ;
\draw [shift={(385.67,169.67)}, rotate = 180] [color={rgb, 255:red, 0; green, 0; blue, 0 }  ][line width=0.75]    (10.93,-3.29) .. controls (6.95,-1.4) and (3.31,-0.3) .. (0,0) .. controls (3.31,0.3) and (6.95,1.4) .. (10.93,3.29)   ;
\draw  [fill={rgb, 255:red, 126; green, 211; blue, 33 }  ,fill opacity=1 ] (496.83,292.41) -- (456.82,280.42) -- (456.81,234.42) -- (496.8,222.41) -- cycle ;
\draw  [fill={rgb, 255:red, 126; green, 211; blue, 33 }  ,fill opacity=1 ] (497.15,205.41) -- (457.14,193.42) -- (457.13,147.42) -- (497.12,135.41) -- cycle ;
\draw  [fill={rgb, 255:red, 126; green, 211; blue, 33 }  ,fill opacity=1 ] (497.48,115.41) -- (457.47,103.42) -- (457.46,57.42) -- (497.45,45.41) -- cycle ;
\draw    (407.67,171.67) -- (454.33,171.67) ;
\draw [shift={(456.33,171.67)}, rotate = 180] [color={rgb, 255:red, 0; green, 0; blue, 0 }  ][line width=0.75]    (10.93,-3.29) .. controls (6.95,-1.4) and (3.31,-0.3) .. (0,0) .. controls (3.31,0.3) and (6.95,1.4) .. (10.93,3.29)   ;
\draw    (407.67,171.67) -- (457.38,79.43) ;
\draw [shift={(458.33,77.67)}, rotate = 118.32] [color={rgb, 255:red, 0; green, 0; blue, 0 }  ][line width=0.75]    (10.93,-3.29) .. controls (6.95,-1.4) and (3.31,-0.3) .. (0,0) .. controls (3.31,0.3) and (6.95,1.4) .. (10.93,3.29)   ;
\draw    (407.67,171.67) -- (455.38,259.91) ;
\draw [shift={(456.33,261.67)}, rotate = 241.6] [color={rgb, 255:red, 0; green, 0; blue, 0 }  ][line width=0.75]    (10.93,-3.29) .. controls (6.95,-1.4) and (3.31,-0.3) .. (0,0) .. controls (3.31,0.3) and (6.95,1.4) .. (10.93,3.29)   ;
\draw  [fill={rgb, 255:red, 80; green, 227; blue, 194 }  ,fill opacity=1 ] (252.33,304.33) -- (375,304.33) -- (375,347.67) -- (252.33,347.67) -- cycle ;
\draw    (312,236) -- (312.97,303.67) ;
\draw [shift={(313,305.67)}, rotate = 269.18] [color={rgb, 255:red, 0; green, 0; blue, 0 }  ][line width=0.75]    (10.93,-3.29) .. controls (6.95,-1.4) and (3.31,-0.3) .. (0,0) .. controls (3.31,0.3) and (6.95,1.4) .. (10.93,3.29)   ;
\draw   (318.33,39.67) .. controls (318.31,35) and (315.97,32.68) .. (311.3,32.7) -- (247.97,32.96) .. controls (241.3,32.99) and (237.96,30.67) .. (237.94,26) .. controls (237.96,30.67) and (234.64,33.02) .. (227.97,33.05)(230.97,33.03) -- (164.64,33.31) .. controls (159.97,33.33) and (157.65,35.67) .. (157.67,40.34) ;
\draw   (499,38.67) .. controls (498.98,34) and (496.64,31.68) .. (491.97,31.7) -- (428.64,31.96) .. controls (421.97,31.99) and (418.63,29.67) .. (418.61,25) .. controls (418.63,29.67) and (415.31,32.02) .. (408.64,32.05)(411.64,32.03) -- (345.3,32.31) .. controls (340.63,32.33) and (338.31,34.67) .. (338.33,39.34) ;

\draw (65.33,154.33) node [anchor=north west][inner sep=0.75pt]   [align=left] {\begin{minipage}[lt]{55.93pt}\setlength\topsep{0pt}
\begin{center}
{\fontfamily{pcr}\selectfont DNA}\\{\fontfamily{pcr}\selectfont methylation}
\end{center}

\end{minipage}};
\draw (69.33,66.67) node [anchor=north west][inner sep=0.75pt]   [align=left] {\begin{minipage}[lt]{48.43pt}\setlength\topsep{0pt}
\begin{center}
{\fontfamily{pcr}\selectfont Gene}\\{\fontfamily{pcr}\selectfont expression}
\end{center}

\end{minipage}};
\draw (68.33,244) node [anchor=north west][inner sep=0.75pt]   [align=left] {\begin{minipage}[lt]{48.43pt}\setlength\topsep{0pt}
\begin{center}
{\fontfamily{pcr}\selectfont miRNA}\\{\fontfamily{pcr}\selectfont expression}
\end{center}

\end{minipage}};
\draw (265.54,309.54) node [anchor=north west][inner sep=0.75pt]  [font=\small,rotate=-0.21] [align=left] {\begin{minipage}[lt]{68.2pt}\setlength\topsep{0pt}
\begin{center}
{\small {\fontfamily{pcr}\selectfont Classification task}}
\end{center}

\end{minipage}};
\draw (305.2,126.87) node [anchor=north west][inner sep=0.75pt]  [font=\footnotesize,rotate=-0.21] [align=left] {\begin{minipage}[lt]{8.67pt}\setlength\topsep{0pt}
\begin{center}
$\displaystyle \boldsymbol{\mu }$
\end{center}

\end{minipage}};
\draw (304.87,204.54) node [anchor=north west][inner sep=0.75pt]  [font=\footnotesize,rotate=-0.21] [align=left] {\begin{minipage}[lt]{8.67pt}\setlength\topsep{0pt}
\begin{center}
$\displaystyle \boldsymbol{\sigma }$
\end{center}

\end{minipage}};
\draw (210.67,6.67) node [anchor=north west][inner sep=0.75pt]   [align=left] {{\fontfamily{pcr}\selectfont Encoder}};
\draw (390,6.33) node [anchor=north west][inner sep=0.75pt]   [align=left] {{\fontfamily{pcr}\selectfont Decoder}};
\draw (509.33,145.67) node [anchor=north west][inner sep=0.75pt]   [align=left] {\begin{minipage}[lt]{55.93pt}\setlength\topsep{0pt}
\begin{center}
{\fontfamily{pcr}\selectfont DNA}\\{\fontfamily{pcr}\selectfont methylation}
\end{center}

\end{minipage}};
\draw (513.33,58.67) node [anchor=north west][inner sep=0.75pt]   [align=left] {\begin{minipage}[lt]{48.43pt}\setlength\topsep{0pt}
\begin{center}
{\fontfamily{pcr}\selectfont Gene}\\{\fontfamily{pcr}\selectfont expression}
\end{center}

\end{minipage}};
\draw (512.33,235.33) node [anchor=north west][inner sep=0.75pt]   [align=left] {\begin{minipage}[lt]{48.43pt}\setlength\topsep{0pt}
\begin{center}
{\fontfamily{pcr}\selectfont miRNA}\\{\fontfamily{pcr}\selectfont expression}
\end{center}

\end{minipage}};
\draw (343.2,165.87) node [anchor=north west][inner sep=0.75pt]  [font=\footnotesize,rotate=-0.21] [align=left] {\begin{minipage}[lt]{8.67pt}\setlength\topsep{0pt}
\begin{center}
$\displaystyle \mathbf{z}$
\end{center}

\end{minipage}};

\end{tikzpicture}

\caption{SubOmiEmbed Architecture: It consists of an encoder that learns meaningful representations of the inputs and a decoder that reconstructs the data from the latent representation. $\mathbf{z}$, $\boldsymbol{\sigma}$ and $\boldsymbol{\mu}$ represent the latent vector, standard deviation vector and mean vector respectively. $\mathbf{z}$ is sampled from $\boldsymbol{\sigma}$ and $\boldsymbol{\mu}$ using the reparameterisation trick shown in Equation \ref{eq:rep}} \label{fig:arch}
\end{figure}
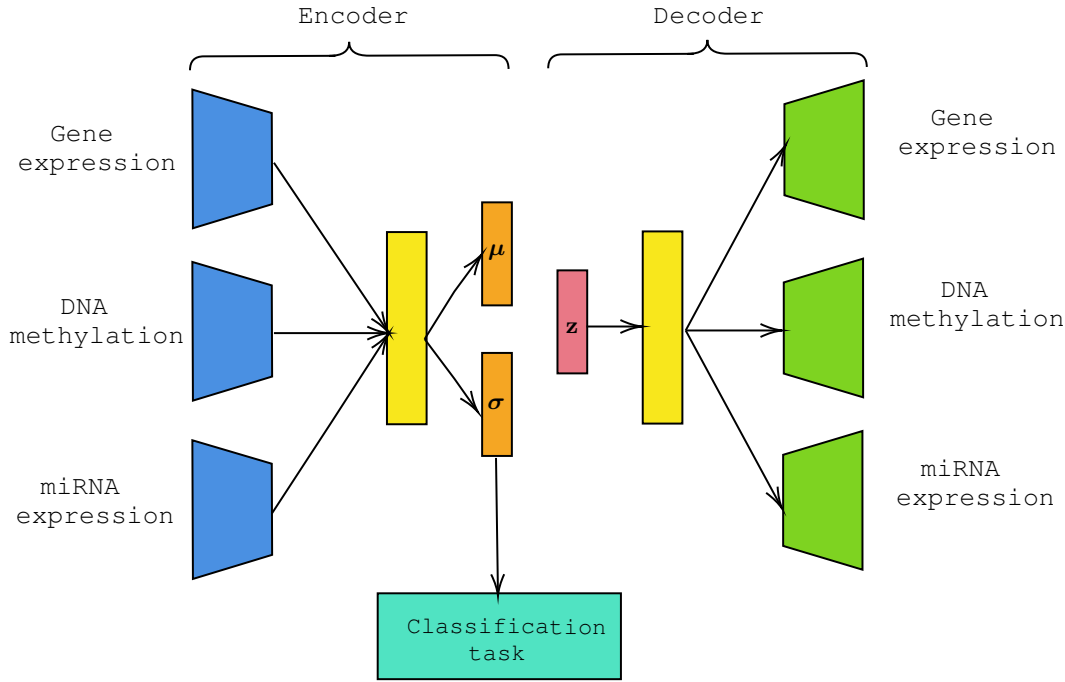

SubOmiEmbed comprises of a deep embedding network and a downstream classification network, as shown in Figure \ref{fig:arch}.
The deep embedding network learns low-dimensional latent space representation from high-dimensional multi-omics data. This latent space is then used for the downstream task of cancer type classification. Fully connected layers are used in the encoder, decoder and downstream network.

The deep embedding module of SubOmiEmbed has a variational autoencoder (VAE) [26] as its backbone. VAE is a generative network that can effectively represent high-dimensional data in a lower dimension. We make the assumption that a latent vector $\mathbf{z}^i \in \mathbb{R}^q$ can be used to represent each sample $\mathbf{x}^i \in \mathbb{R}^d$ in the dataset $\mathcal{D}$, where $q \gg d$. To generate samples, each latent vector is sampled from a prior distribution $p_\theta(\mathbf{z})$, after which the conditional distribution $p_\theta(\mathbf{x} \mid \mathbf{z})$ generates multi-omics data for each sample. Here, $\theta$ is the set of decoder parameters that can be learned. Due to the difficulty in estimating the true posterior $p_\theta(\mathbf{z} \mid \mathbf{x})$, the variational distribution $q_\phi(\mathbf{z} \mid \mathbf{x})$ is used to approximate $p_\theta(\mathbf{z} \mid \mathbf{x})$. Here, $\phi$ is the set of encoder parameters that can be learned. Therefore, by maximizing the variational lower bound given below, the VAE model can be optimised:
\begin{equation}
\mathbb{E}_{\mathbf{z} \sim q_{\phi}(\mathbf{z} \mid \mathbf{x})} \log p_{\theta}(\mathbf{x} \mid \mathbf{z})-D_{\mathrm{KL}}\left(q_{\phi}(\mathbf{z} \mid \mathbf{x}) \| p_{\theta}(\mathbf{z})\right)
\end{equation}
Here, $D_{\mathrm{KL}}$is the Kullback–Leibler (KL) divergence between 2 probability distributions \cite{dl}.

Each of the three types of input data was first encoded into respective vectors. These vectors were then concatenated together before feeding them into the hidden layers in the encoder. To produce the standard deviation vector $\bm{\sigma}$ and the mean vector $\bm{\mu}$, the output vector from hidden layers was connected to 2 bottleneck layers. The variational distribution $q_\phi(\mathbf{z} \mid \mathbf{x})$ is defined by these vectors, which is also the Gaussian distribution $\mathcal{N}(\bm{\mu}, \bm{\sigma})$ of the latent variable $\mathbf{z}$ given the input $\mathbf{x}$. The reparameterization trick as shown in equation \ref{eq:rep}, is used because sampling $\mathbf{z}$ from the distribution is not differentiable and is unsuitable for backpropagation. 
\begin{equation}
\label{eq:rep}
    \mathbf{z} = \bm{\mu} + \bm{\sigma} \bm{\epsilon}
\end{equation}
Here, $\bm{\epsilon}$ is a random variable which is sampled from the standard Gaussian distribution $\mathcal{N}(\bm{0}, \bm{I})$. To reconstruct the input data $\mathbf{x'}$, latent variable $\mathbf{z}$ is then passed to a symmetrical decoder.

\subsection{Training Strategy}
The loss function in the model comprises of two major parts, the embedding loss and classification loss. Let's denote the input data $\mathbf{x}$ of $i$th omics type as $\mathbf{x}_i$ and the reconstructed data as $\mathbf{x}'_i$. Now the embedding loss can be formulated as shown below.
\begin{equation}
\mathcal{L}_{\text {embed }}=\frac{1}{N} \sum_{i=1}^{N} MSE\left(\mathbf{x}_{i}, \mathbf{x}_{i}^{\prime}\right)+D_{\mathrm{KL}}(\mathcal{N}(\boldsymbol{\mu}, \boldsymbol{\sigma}) \| \mathcal{N}(\mathbf{0}, \mathbf{I}))
\end{equation}
Here, the 1st part of the loss is the mean squared error loss between input and reconstructed omics data, while the 2nd part represents the KL divergence between the learnt probability distribution and a standard normal distribution.

Loss function for the downstream classification task is formulated as follows
\begin{equation}
    \mathcal{L}_{\text {classification}} =  CE(y, {y}')
    \label{eq:class}
\end{equation}
Here, $y$ is the label, ${y}'$ the prediction and CE the cross entropy loss.

Additionally, the joint loss is defined as
\begin{equation}
\label{eq:joint}
     \mathcal{L}_{\text {joint}} = \mathcal{L}_{\text {embed}} + \mathcal{L}_{\text {classification}}
\end{equation}

Training is divided into three phases. The 1st phase is unsupervised where the loss minimized is the embedding loss and the encoder and decoder parameters are learnt. Labels are not needed in this phase and phase 1 could be used individually for visualisation or reducing dimensions. In phase 2, the layers of embedding module are kept frozen while the layers of downstream network are trained. Here, only the classification loss is minimized. 
In the third phase, the overall loss function as shown in Equation \ref{eq:joint} is minimized. In this last phase, the whole network consisting of both the embedding layers and the classification network layers are trained. This way, the network learns representations that are useful for the downstream classification task.

\subsection{Feature subsetting}
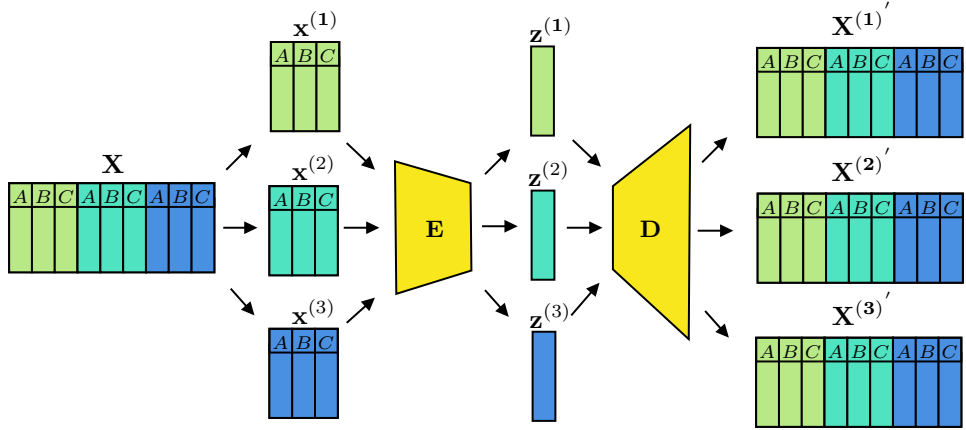
\begin{figure}[h]
    \centering

\tikzset{every picture/.style={line width=0.75pt}} 

\begin{tikzpicture}[x=0.75pt,y=0.75pt,yscale=-0.75,xscale=0.75]

\draw  [fill={rgb, 255:red, 184; green, 233; blue, 134 }  ,fill opacity=1 ] (511.33,242.41) -- (557.33,242.41) -- (557.33,292) -- (511.33,292) -- cycle ;
\draw  [fill={rgb, 255:red, 184; green, 233; blue, 134 }  ,fill opacity=1 ] (511.33,232) -- (526.67,232) -- (526.67,292) -- (511.33,292) -- cycle ;
\draw  [fill={rgb, 255:red, 184; green, 233; blue, 134 }  ,fill opacity=1 ] (526.67,232) -- (542,232) -- (542,292) -- (526.67,292) -- cycle ;
\draw  [fill={rgb, 255:red, 184; green, 233; blue, 134 }  ,fill opacity=1 ] (542,232) -- (557.33,232) -- (557.33,292) -- (542,292) -- cycle ;
\draw   (511.33,232) -- (557.33,232) -- (557.33,248) -- (511.33,248) -- cycle ;
\draw  [fill={rgb, 255:red, 80; green, 227; blue, 194 }  ,fill opacity=1 ] (557.33,242.41) -- (603.33,242.41) -- (603.33,292) -- (557.33,292) -- cycle ;
\draw  [fill={rgb, 255:red, 80; green, 227; blue, 194 }  ,fill opacity=1 ] (557.33,232) -- (572.67,232) -- (572.67,292) -- (557.33,292) -- cycle ;
\draw  [fill={rgb, 255:red, 80; green, 227; blue, 194 }  ,fill opacity=1 ] (572.67,232) -- (588,232) -- (588,292) -- (572.67,292) -- cycle ;
\draw  [fill={rgb, 255:red, 80; green, 227; blue, 194 }  ,fill opacity=1 ] (588,232) -- (603.33,232) -- (603.33,292) -- (588,292) -- cycle ;
\draw   (557.33,232) -- (603.33,232) -- (603.33,248) -- (557.33,248) -- cycle ;
\draw  [fill={rgb, 255:red, 74; green, 144; blue, 226 }  ,fill opacity=1 ] (603.33,242.41) -- (649.33,242.41) -- (649.33,292) -- (603.33,292) -- cycle ;
\draw  [fill={rgb, 255:red, 74; green, 144; blue, 226 }  ,fill opacity=1 ] (603.33,232) -- (618.67,232) -- (618.67,292) -- (603.33,292) -- cycle ;
\draw  [fill={rgb, 255:red, 74; green, 144; blue, 226 }  ,fill opacity=1 ] (618.67,232) -- (634,232) -- (634,292) -- (618.67,292) -- cycle ;
\draw  [fill={rgb, 255:red, 74; green, 144; blue, 226 }  ,fill opacity=1 ] (634,232) -- (649.33,232) -- (649.33,292) -- (634,292) -- cycle ;
\draw   (603.33,232) -- (649.33,232) -- (649.33,248) -- (603.33,248) -- cycle ;
\draw  [fill={rgb, 255:red, 184; green, 233; blue, 134 }  ,fill opacity=1 ] (512.33,145.41) -- (558.33,145.41) -- (558.33,195) -- (512.33,195) -- cycle ;
\draw  [fill={rgb, 255:red, 184; green, 233; blue, 134 }  ,fill opacity=1 ] (512.33,135) -- (527.67,135) -- (527.67,195) -- (512.33,195) -- cycle ;
\draw  [fill={rgb, 255:red, 184; green, 233; blue, 134 }  ,fill opacity=1 ] (527.67,135) -- (543,135) -- (543,195) -- (527.67,195) -- cycle ;
\draw  [fill={rgb, 255:red, 184; green, 233; blue, 134 }  ,fill opacity=1 ] (543,135) -- (558.33,135) -- (558.33,195) -- (543,195) -- cycle ;
\draw   (512.33,135) -- (558.33,135) -- (558.33,151) -- (512.33,151) -- cycle ;
\draw  [fill={rgb, 255:red, 80; green, 227; blue, 194 }  ,fill opacity=1 ] (558.33,145.41) -- (604.33,145.41) -- (604.33,195) -- (558.33,195) -- cycle ;
\draw  [fill={rgb, 255:red, 80; green, 227; blue, 194 }  ,fill opacity=1 ] (558.33,135) -- (573.67,135) -- (573.67,195) -- (558.33,195) -- cycle ;
\draw  [fill={rgb, 255:red, 80; green, 227; blue, 194 }  ,fill opacity=1 ] (573.67,135) -- (589,135) -- (589,195) -- (573.67,195) -- cycle ;
\draw  [fill={rgb, 255:red, 80; green, 227; blue, 194 }  ,fill opacity=1 ] (589,135) -- (604.33,135) -- (604.33,195) -- (589,195) -- cycle ;
\draw   (558.33,135) -- (604.33,135) -- (604.33,151) -- (558.33,151) -- cycle ;
\draw  [fill={rgb, 255:red, 74; green, 144; blue, 226 }  ,fill opacity=1 ] (604.33,145.41) -- (650.33,145.41) -- (650.33,195) -- (604.33,195) -- cycle ;
\draw  [fill={rgb, 255:red, 74; green, 144; blue, 226 }  ,fill opacity=1 ] (604.33,135) -- (619.67,135) -- (619.67,195) -- (604.33,195) -- cycle ;
\draw  [fill={rgb, 255:red, 74; green, 144; blue, 226 }  ,fill opacity=1 ] (619.67,135) -- (635,135) -- (635,195) -- (619.67,195) -- cycle ;
\draw  [fill={rgb, 255:red, 74; green, 144; blue, 226 }  ,fill opacity=1 ] (635,135) -- (650.33,135) -- (650.33,195) -- (635,195) -- cycle ;
\draw   (604.33,135) -- (650.33,135) -- (650.33,151) -- (604.33,151) -- cycle ;
\draw  [fill={rgb, 255:red, 184; green, 233; blue, 134 }  ,fill opacity=1 ] (512,47.41) -- (558,47.41) -- (558,97) -- (512,97) -- cycle ;
\draw  [fill={rgb, 255:red, 184; green, 233; blue, 134 }  ,fill opacity=1 ] (512,37) -- (527.33,37) -- (527.33,97) -- (512,97) -- cycle ;
\draw  [fill={rgb, 255:red, 184; green, 233; blue, 134 }  ,fill opacity=1 ] (527.33,37) -- (542.67,37) -- (542.67,97) -- (527.33,97) -- cycle ;
\draw  [fill={rgb, 255:red, 184; green, 233; blue, 134 }  ,fill opacity=1 ] (542.67,37) -- (558,37) -- (558,97) -- (542.67,97) -- cycle ;
\draw   (512,37) -- (558,37) -- (558,53) -- (512,53) -- cycle ;
\draw  [fill={rgb, 255:red, 80; green, 227; blue, 194 }  ,fill opacity=1 ] (558,47.41) -- (604,47.41) -- (604,97) -- (558,97) -- cycle ;
\draw  [fill={rgb, 255:red, 80; green, 227; blue, 194 }  ,fill opacity=1 ] (558,37) -- (573.33,37) -- (573.33,97) -- (558,97) -- cycle ;
\draw  [fill={rgb, 255:red, 80; green, 227; blue, 194 }  ,fill opacity=1 ] (573.33,37) -- (588.67,37) -- (588.67,97) -- (573.33,97) -- cycle ;
\draw  [fill={rgb, 255:red, 80; green, 227; blue, 194 }  ,fill opacity=1 ] (588.67,37) -- (604,37) -- (604,97) -- (588.67,97) -- cycle ;
\draw   (558,37) -- (604,37) -- (604,53) -- (558,53) -- cycle ;
\draw  [fill={rgb, 255:red, 74; green, 144; blue, 226 }  ,fill opacity=1 ] (604,47.41) -- (650,47.41) -- (650,97) -- (604,97) -- cycle ;
\draw  [fill={rgb, 255:red, 74; green, 144; blue, 226 }  ,fill opacity=1 ] (604,37) -- (619.33,37) -- (619.33,97) -- (604,97) -- cycle ;
\draw  [fill={rgb, 255:red, 74; green, 144; blue, 226 }  ,fill opacity=1 ] (619.33,37) -- (634.67,37) -- (634.67,97) -- (619.33,97) -- cycle ;
\draw  [fill={rgb, 255:red, 74; green, 144; blue, 226 }  ,fill opacity=1 ] (634.67,37) -- (650,37) -- (650,97) -- (634.67,97) -- cycle ;
\draw   (604,37) -- (650,37) -- (650,53) -- (604,53) -- cycle ;
\draw  [fill={rgb, 255:red, 74; green, 144; blue, 226 }  ,fill opacity=1 ] (184,236.41) -- (230,236.41) -- (230,286) -- (184,286) -- cycle ;
\draw  [fill={rgb, 255:red, 74; green, 144; blue, 226 }  ,fill opacity=1 ] (184,226) -- (199.33,226) -- (199.33,286) -- (184,286) -- cycle ;
\draw  [fill={rgb, 255:red, 74; green, 144; blue, 226 }  ,fill opacity=1 ] (199.33,226) -- (214.67,226) -- (214.67,286) -- (199.33,286) -- cycle ;
\draw  [fill={rgb, 255:red, 74; green, 144; blue, 226 }  ,fill opacity=1 ] (214.67,226) -- (230,226) -- (230,286) -- (214.67,286) -- cycle ;
\draw   (184,226) -- (230,226) -- (230,242) -- (184,242) -- cycle ;
\draw  [fill={rgb, 255:red, 80; green, 227; blue, 194 }  ,fill opacity=1 ] (184,140.41) -- (230,140.41) -- (230,190) -- (184,190) -- cycle ;
\draw  [fill={rgb, 255:red, 80; green, 227; blue, 194 }  ,fill opacity=1 ] (184,130) -- (199.33,130) -- (199.33,190) -- (184,190) -- cycle ;
\draw  [fill={rgb, 255:red, 80; green, 227; blue, 194 }  ,fill opacity=1 ] (199.33,130) -- (214.67,130) -- (214.67,190) -- (199.33,190) -- cycle ;
\draw  [fill={rgb, 255:red, 80; green, 227; blue, 194 }  ,fill opacity=1 ] (214.67,130) -- (230,130) -- (230,190) -- (214.67,190) -- cycle ;
\draw   (184,130) -- (230,130) -- (230,146) -- (184,146) -- cycle ;
\draw  [fill={rgb, 255:red, 184; green, 233; blue, 134 }  ,fill opacity=1 ] (185,43.41) -- (231,43.41) -- (231,93) -- (185,93) -- cycle ;
\draw  [fill={rgb, 255:red, 184; green, 233; blue, 134 }  ,fill opacity=1 ] (185,33) -- (200.33,33) -- (200.33,93) -- (185,93) -- cycle ;
\draw  [fill={rgb, 255:red, 184; green, 233; blue, 134 }  ,fill opacity=1 ] (200.33,33) -- (215.67,33) -- (215.67,93) -- (200.33,93) -- cycle ;
\draw  [fill={rgb, 255:red, 184; green, 233; blue, 134 }  ,fill opacity=1 ] (215.67,33) -- (231,33) -- (231,93) -- (215.67,93) -- cycle ;
\draw   (185,33) -- (231,33) -- (231,49) -- (185,49) -- cycle ;
\draw  [fill={rgb, 255:red, 248; green, 231; blue, 28 }  ,fill opacity=1 ] (268.45,113.31) -- (319.26,128) -- (319.85,186.8) -- (269.34,202.5) -- cycle ;
\draw  [fill={rgb, 255:red, 184; green, 233; blue, 134 }  ,fill opacity=1 ] (9,138.41) -- (55,138.41) -- (55,188) -- (9,188) -- cycle ;
\draw  [fill={rgb, 255:red, 184; green, 233; blue, 134 }  ,fill opacity=1 ] (9,128) -- (24.33,128) -- (24.33,188) -- (9,188) -- cycle ;
\draw  [fill={rgb, 255:red, 184; green, 233; blue, 134 }  ,fill opacity=1 ] (24.33,128) -- (39.67,128) -- (39.67,188) -- (24.33,188) -- cycle ;
\draw  [fill={rgb, 255:red, 184; green, 233; blue, 134 }  ,fill opacity=1 ] (39.67,128) -- (55,128) -- (55,188) -- (39.67,188) -- cycle ;
\draw   (9,128) -- (55,128) -- (55,144) -- (9,144) -- cycle ;
\draw  [fill={rgb, 255:red, 80; green, 227; blue, 194 }  ,fill opacity=1 ] (55,138.41) -- (101,138.41) -- (101,188) -- (55,188) -- cycle ;
\draw  [fill={rgb, 255:red, 80; green, 227; blue, 194 }  ,fill opacity=1 ] (55,128) -- (70.33,128) -- (70.33,188) -- (55,188) -- cycle ;
\draw  [fill={rgb, 255:red, 80; green, 227; blue, 194 }  ,fill opacity=1 ] (70.33,128) -- (85.67,128) -- (85.67,188) -- (70.33,188) -- cycle ;
\draw  [fill={rgb, 255:red, 80; green, 227; blue, 194 }  ,fill opacity=1 ] (85.67,128) -- (101,128) -- (101,188) -- (85.67,188) -- cycle ;
\draw   (55,128) -- (101,128) -- (101,144) -- (55,144) -- cycle ;
\draw  [fill={rgb, 255:red, 74; green, 144; blue, 226 }  ,fill opacity=1 ] (101,138.41) -- (147,138.41) -- (147,188) -- (101,188) -- cycle ;
\draw  [fill={rgb, 255:red, 74; green, 144; blue, 226 }  ,fill opacity=1 ] (101,128) -- (116.33,128) -- (116.33,188) -- (101,188) -- cycle ;
\draw  [fill={rgb, 255:red, 74; green, 144; blue, 226 }  ,fill opacity=1 ] (116.33,128) -- (131.67,128) -- (131.67,188) -- (116.33,188) -- cycle ;
\draw  [fill={rgb, 255:red, 74; green, 144; blue, 226 }  ,fill opacity=1 ] (131.67,128) -- (147,128) -- (147,188) -- (131.67,188) -- cycle ;
\draw   (101,128) -- (147,128) -- (147,144) -- (101,144) -- cycle ;
\draw  [fill={rgb, 255:red, 184; green, 233; blue, 134 }  ,fill opacity=1 ] (360,36) -- (375.33,36) -- (375.33,96) -- (360,96) -- cycle ;
\draw  [fill={rgb, 255:red, 80; green, 227; blue, 194 }  ,fill opacity=1 ] (360.33,133) -- (375.67,133) -- (375.67,193) -- (360.33,193) -- cycle ;
\draw  [fill={rgb, 255:red, 74; green, 144; blue, 226 }  ,fill opacity=1 ] (361,228) -- (376.33,228) -- (376.33,288) -- (361,288) -- cycle ;
\draw    (153,158) -- (173,158) ;
\draw [shift={(176,158)}, rotate = 180] [fill={rgb, 255:red, 0; green, 0; blue, 0 }  ][line width=0.08]  [draw opacity=0] (8.93,-4.29) -- (0,0) -- (8.93,4.29) -- cycle    ;
\draw    (234,158) -- (255,158) ;
\draw [shift={(258,158)}, rotate = 180] [fill={rgb, 255:red, 0; green, 0; blue, 0 }  ][line width=0.08]  [draw opacity=0] (8.93,-4.29) -- (0,0) -- (8.93,4.29) -- cycle    ;
\draw    (327,157) -- (348,157) ;
\draw [shift={(351,157)}, rotate = 180] [fill={rgb, 255:red, 0; green, 0; blue, 0 }  ][line width=0.08]  [draw opacity=0] (8.93,-4.29) -- (0,0) -- (8.93,4.29) -- cycle    ;
\draw    (384,158) -- (405,158) ;
\draw [shift={(408,158)}, rotate = 180] [fill={rgb, 255:red, 0; green, 0; blue, 0 }  ][line width=0.08]  [draw opacity=0] (8.93,-4.29) -- (0,0) -- (8.93,4.29) -- cycle    ;
\draw    (236,101) -- (252.82,116.94) ;
\draw [shift={(255,119)}, rotate = 223.45] [fill={rgb, 255:red, 0; green, 0; blue, 0 }  ][line width=0.08]  [draw opacity=0] (8.93,-4.29) -- (0,0) -- (8.93,4.29) -- cycle    ;
\draw    (388,99) -- (404.82,114.94) ;
\draw [shift={(407,117)}, rotate = 223.45] [fill={rgb, 255:red, 0; green, 0; blue, 0 }  ][line width=0.08]  [draw opacity=0] (8.93,-4.29) -- (0,0) -- (8.93,4.29) -- cycle    ;
\draw    (236,222) -- (252.76,207) ;
\draw [shift={(255,205)}, rotate = 138.18] [fill={rgb, 255:red, 0; green, 0; blue, 0 }  ][line width=0.08]  [draw opacity=0] (8.93,-4.29) -- (0,0) -- (8.93,4.29) -- cycle    ;
\draw    (387,217) -- (404.04,197.27) ;
\draw [shift={(406,195)}, rotate = 130.82] [fill={rgb, 255:red, 0; green, 0; blue, 0 }  ][line width=0.08]  [draw opacity=0] (8.93,-4.29) -- (0,0) -- (8.93,4.29) -- cycle    ;
\draw    (329,122) -- (343.88,107.12) ;
\draw [shift={(346,105)}, rotate = 135] [fill={rgb, 255:red, 0; green, 0; blue, 0 }  ][line width=0.08]  [draw opacity=0] (8.93,-4.29) -- (0,0) -- (8.93,4.29) -- cycle    ;
\draw    (329,200) -- (343.01,215.76) ;
\draw [shift={(345,218)}, rotate = 228.37] [fill={rgb, 255:red, 0; green, 0; blue, 0 }  ][line width=0.08]  [draw opacity=0] (8.93,-4.29) -- (0,0) -- (8.93,4.29) -- cycle    ;
\draw    (155,119) -- (169.88,104.12) ;
\draw [shift={(172,102)}, rotate = 135] [fill={rgb, 255:red, 0; green, 0; blue, 0 }  ][line width=0.08]  [draw opacity=0] (8.93,-4.29) -- (0,0) -- (8.93,4.29) -- cycle    ;
\draw    (158,199) -- (172.01,214.76) ;
\draw [shift={(174,217)}, rotate = 228.37] [fill={rgb, 255:red, 0; green, 0; blue, 0 }  ][line width=0.08]  [draw opacity=0] (8.93,-4.29) -- (0,0) -- (8.93,4.29) -- cycle    ;
\draw  [fill={rgb, 255:red, 248; green, 231; blue, 28 }  ,fill opacity=1 ] (415,130) -- (467,89) -- (466,233) -- (415,186) -- cycle ;
\draw    (472,160) -- (493,160) ;
\draw [shift={(496,160)}, rotate = 180] [fill={rgb, 255:red, 0; green, 0; blue, 0 }  ][line width=0.08]  [draw opacity=0] (8.93,-4.29) -- (0,0) -- (8.93,4.29) -- cycle    ;
\draw    (475,114) -- (489.88,99.12) ;
\draw [shift={(492,97)}, rotate = 135] [fill={rgb, 255:red, 0; green, 0; blue, 0 }  ][line width=0.08]  [draw opacity=0] (8.93,-4.29) -- (0,0) -- (8.93,4.29) -- cycle    ;
\draw    (477,214) -- (491.01,229.76) ;
\draw [shift={(493,232)}, rotate = 228.37] [fill={rgb, 255:red, 0; green, 0; blue, 0 }  ][line width=0.08]  [draw opacity=0] (8.93,-4.29) -- (0,0) -- (8.93,4.29) -- cycle    ;

\draw (69,107) node [anchor=north west][inner sep=0.75pt]   [align=left] {$\displaystyle \mathbf{X}$};
\draw (198,10) node [anchor=north west][inner sep=0.75pt]   [align=left] {$\displaystyle \mathbf{x^{( 1)}}$};
\draw (197,107) node [anchor=north west][inner sep=0.75pt]   [align=left] {$\displaystyle \mathbf{x}^{( 2)}$};
\draw (197,204) node [anchor=north west][inner sep=0.75pt]   [align=left] {$\displaystyle \mathbf{x}^{( 3)}$};
\draw (357,14) node [anchor=north west][inner sep=0.75pt]   [align=left] {$\displaystyle \mathbf{z^{( 1)}}$};
\draw (356,111) node [anchor=north west][inner sep=0.75pt]   [align=left] {$\displaystyle \mathbf{z}^{( 2)}$};
\draw (357,207) node [anchor=north west][inner sep=0.75pt]   [align=left] {$\displaystyle \mathbf{z}^{( 3)}$};
\draw (560,4) node [anchor=north west][inner sep=0.75pt]   [align=left] {$\displaystyle \mathbf{X{^{( 1)}}}^{'}$};
\draw (287,151) node [anchor=north west][inner sep=0.75pt]  [font=\small] [align=left] {$\displaystyle \mathbf{E}$};
\draw (431,151) node [anchor=north west][inner sep=0.75pt]  [font=\small] [align=left] {$\displaystyle \mathbf{D}$};
\draw (512,40) node [anchor=north west][inner sep=0.75pt]  [font=\scriptsize] [align=left] {$\displaystyle A$};
\draw (542.67,40) node [anchor=north west][inner sep=0.75pt]  [font=\scriptsize] [align=left] {$\displaystyle C$};
\draw (588.67,40) node [anchor=north west][inner sep=0.75pt]  [font=\scriptsize] [align=left] {$\displaystyle C$};
\draw (558,40) node [anchor=north west][inner sep=0.75pt]  [font=\scriptsize] [align=left] {$\displaystyle A$};
\draw (604,40) node [anchor=north west][inner sep=0.75pt]  [font=\scriptsize] [align=left] {$\displaystyle A$};
\draw (634.67,40) node [anchor=north west][inner sep=0.75pt]  [font=\scriptsize] [align=left] {$\displaystyle C$};
\draw (512.33,138) node [anchor=north west][inner sep=0.75pt]  [font=\scriptsize] [align=left] {$\displaystyle A$};
\draw (543,138) node [anchor=north west][inner sep=0.75pt]  [font=\scriptsize] [align=left] {$\displaystyle C$};
\draw (589,138) node [anchor=north west][inner sep=0.75pt]  [font=\scriptsize] [align=left] {$\displaystyle C$};
\draw (558.33,138) node [anchor=north west][inner sep=0.75pt]  [font=\scriptsize] [align=left] {$\displaystyle A$};
\draw (604.33,138) node [anchor=north west][inner sep=0.75pt]  [font=\scriptsize] [align=left] {$\displaystyle A$};
\draw (635,138) node [anchor=north west][inner sep=0.75pt]  [font=\scriptsize] [align=left] {$\displaystyle C$};
\draw (511.33,235) node [anchor=north west][inner sep=0.75pt]  [font=\scriptsize] [align=left] {$\displaystyle A$};
\draw (542,235) node [anchor=north west][inner sep=0.75pt]  [font=\scriptsize] [align=left] {$\displaystyle C$};
\draw (588,235) node [anchor=north west][inner sep=0.75pt]  [font=\scriptsize] [align=left] {$\displaystyle C$};
\draw (557.33,235) node [anchor=north west][inner sep=0.75pt]  [font=\scriptsize] [align=left] {$\displaystyle A$};
\draw (603.33,235) node [anchor=north west][inner sep=0.75pt]  [font=\scriptsize] [align=left] {$\displaystyle A$};
\draw (634,235) node [anchor=north west][inner sep=0.75pt]  [font=\scriptsize] [align=left] {$\displaystyle C$};
\draw (185,36) node [anchor=north west][inner sep=0.75pt]  [font=\scriptsize] [align=left] {$\displaystyle A$};
\draw (215.67,36) node [anchor=north west][inner sep=0.75pt]  [font=\scriptsize] [align=left] {$\displaystyle C$};
\draw (214.67,133) node [anchor=north west][inner sep=0.75pt]  [font=\scriptsize] [align=left] {$\displaystyle C$};
\draw (184,133) node [anchor=north west][inner sep=0.75pt]  [font=\scriptsize] [align=left] {$\displaystyle A$};
\draw (184,229) node [anchor=north west][inner sep=0.75pt]  [font=\scriptsize] [align=left] {$\displaystyle A$};
\draw (214.67,229) node [anchor=north west][inner sep=0.75pt]  [font=\scriptsize] [align=left] {$\displaystyle C$};
\draw (101,131) node [anchor=north west][inner sep=0.75pt]  [font=\scriptsize] [align=left] {$\displaystyle A$};
\draw (131.67,131) node [anchor=north west][inner sep=0.75pt]  [font=\scriptsize] [align=left] {$\displaystyle C$};
\draw (85.67,131) node [anchor=north west][inner sep=0.75pt]  [font=\scriptsize] [align=left] {$\displaystyle C$};
\draw (55,131) node [anchor=north west][inner sep=0.75pt]  [font=\scriptsize] [align=left] {$\displaystyle A$};
\draw (9,131) node [anchor=north west][inner sep=0.75pt]  [font=\scriptsize] [align=left] {$\displaystyle A$};
\draw (39.67,131) node [anchor=north west][inner sep=0.75pt]  [font=\scriptsize] [align=left] {$\displaystyle C$};
\draw (560,102) node [anchor=north west][inner sep=0.75pt]   [align=left] {$\displaystyle \mathbf{X{^{( 2)}}}^{'}$};
\draw (560.33,199) node [anchor=north west][inner sep=0.75pt]   [align=left] {$\displaystyle \mathbf{X{^{( 3)}}}^{'}$};
\draw (527.33,40) node [anchor=north west][inner sep=0.75pt]  [font=\scriptsize] [align=left] {$\displaystyle B$};
\draw (573.33,40) node [anchor=north west][inner sep=0.75pt]  [font=\scriptsize] [align=left] {$\displaystyle B$};
\draw (619.33,40) node [anchor=north west][inner sep=0.75pt]  [font=\scriptsize] [align=left] {$\displaystyle B$};
\draw (527.67,138) node [anchor=north west][inner sep=0.75pt]  [font=\scriptsize] [align=left] {$\displaystyle B$};
\draw (573.67,138) node [anchor=north west][inner sep=0.75pt]  [font=\scriptsize] [align=left] {$\displaystyle B$};
\draw (619.67,138) node [anchor=north west][inner sep=0.75pt]  [font=\scriptsize] [align=left] {$\displaystyle B$};
\draw (526.67,235) node [anchor=north west][inner sep=0.75pt]  [font=\scriptsize] [align=left] {$\displaystyle B$};
\draw (572.67,235) node [anchor=north west][inner sep=0.75pt]  [font=\scriptsize] [align=left] {$\displaystyle B$};
\draw (618.67,235) node [anchor=north west][inner sep=0.75pt]  [font=\scriptsize] [align=left] {$\displaystyle B$};
\draw (200.33,36) node [anchor=north west][inner sep=0.75pt]  [font=\scriptsize] [align=left] {$\displaystyle B$};
\draw (199.33,133) node [anchor=north west][inner sep=0.75pt]  [font=\scriptsize] [align=left] {$\displaystyle B$};
\draw (199.33,229) node [anchor=north west][inner sep=0.75pt]  [font=\scriptsize] [align=left] {$\displaystyle B$};
\draw (116.33,131) node [anchor=north west][inner sep=0.75pt]  [font=\scriptsize] [align=left] {$\displaystyle B$};
\draw (70.33,131) node [anchor=north west][inner sep=0.75pt]  [font=\scriptsize] [align=left] {$\displaystyle B$};
\draw (24.33,131) node [anchor=north west][inner sep=0.75pt]  [font=\scriptsize] [align=left] {$\displaystyle B$};

\end{tikzpicture}

    \caption{Illustration of feature subsetting during training. Here, the dataset $\mathbf{X}$ is divided into 3 subsets, $\mathbf{x}^{(1)}$, $\mathbf{x}^{(2)}$ and $\mathbf{x}^{(3)}$. All omics-type A, B and C are proportionately divided into 3 subsets. For instance, $\mathbf{x}^{(1)}$ contains $\frac{1}{3}$rd of each type A, B and C. Each subset is fed to the same encoder which produces latent representations $\mathbf{z}^{(1)}$, $\mathbf{z}^{(2)}$ and $\mathbf{z}^{(3)}$ for the corresponding subsets. These representations are then fed into the decoder. The decoder reconstructs the whole training set $\mathbf{X}$. The reconstruction from $\mathbf{z}^{(1)}$ is named ${\mathbf{X}^{(1)}}^{'}$ and so on. }
    \label{fig:subtabtrain}
\end{figure}

In this setting, we divided the features into subsets and trained encoder and decoder on each of the subsets. This is shown in Figure \ref{fig:subtabtrain}. Suppose we divide the $i$th omics-type dataset $\mathbf{x}_i$ into $M$ subsets. For each subset $\mathbf{x}_i^{(j)}$, we reconstruct the entire dataset and the reconstruction is denoted by ${\mathbf{x}_i^{(j)}}'$. The embedding loss for $j$th subset can be computed as shown below
\begin{equation}
{\mathcal{L}_{\text {embed}}}^{(j)}=\frac{1}{N} \sum_{i=1}^{N} MSE\left(\mathbf{x}_{i}^{(j)}, {\mathbf{x}_{i}^{(j)}}^{\prime}\right)+D_{\mathrm{KL}}(\mathcal{N}(\boldsymbol{\mu}^{(j)}, \boldsymbol{\sigma}^{(j)}) \| \mathcal{N}(\mathbf{0}, \mathbf{I}))
\end{equation}

The joint loss is given as
\begin{equation}
\mathcal{L}_\text{joint}=\frac{1}{M} \sum_{j=1}^{M} {\mathcal{L}_{\text {embed}}}^{(j)}+ \mathcal{L}_{\text {classification}}
\end{equation}

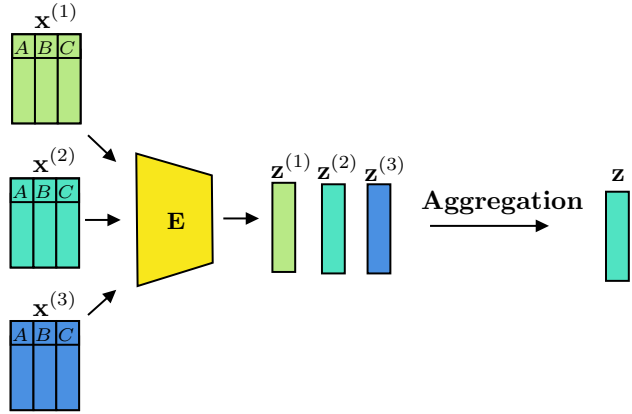
\begin{figure}[h]
    \centering
    \begin{tikzpicture}[x=0.75pt,y=0.75pt,yscale=-0.75,xscale=0.75]

\draw  [fill={rgb, 255:red, 74; green, 144; blue, 226 }  ,fill opacity=1 ] (136,234.41) -- (182,234.41) -- (182,284) -- (136,284) -- cycle ;
\draw  [fill={rgb, 255:red, 74; green, 144; blue, 226 }  ,fill opacity=1 ] (136,224) -- (151.33,224) -- (151.33,284) -- (136,284) -- cycle ;
\draw  [fill={rgb, 255:red, 74; green, 144; blue, 226 }  ,fill opacity=1 ] (151.33,224) -- (166.67,224) -- (166.67,284) -- (151.33,284) -- cycle ;
\draw  [fill={rgb, 255:red, 74; green, 144; blue, 226 }  ,fill opacity=1 ] (166.67,224) -- (182,224) -- (182,284) -- (166.67,284) -- cycle ;
\draw   (136,224) -- (182,224) -- (182,240) -- (136,240) -- cycle ;
\draw  [fill={rgb, 255:red, 80; green, 227; blue, 194 }  ,fill opacity=1 ] (136,138.41) -- (182,138.41) -- (182,188) -- (136,188) -- cycle ;
\draw  [fill={rgb, 255:red, 80; green, 227; blue, 194 }  ,fill opacity=1 ] (136,128) -- (151.33,128) -- (151.33,188) -- (136,188) -- cycle ;
\draw  [fill={rgb, 255:red, 80; green, 227; blue, 194 }  ,fill opacity=1 ] (151.33,128) -- (166.67,128) -- (166.67,188) -- (151.33,188) -- cycle ;
\draw  [fill={rgb, 255:red, 80; green, 227; blue, 194 }  ,fill opacity=1 ] (166.67,128) -- (182,128) -- (182,188) -- (166.67,188) -- cycle ;
\draw   (136,128) -- (182,128) -- (182,144) -- (136,144) -- cycle ;
\draw  [fill={rgb, 255:red, 184; green, 233; blue, 134 }  ,fill opacity=1 ] (137,41.41) -- (183,41.41) -- (183,91) -- (137,91) -- cycle ;
\draw  [fill={rgb, 255:red, 184; green, 233; blue, 134 }  ,fill opacity=1 ] (137,31) -- (152.33,31) -- (152.33,91) -- (137,91) -- cycle ;
\draw  [fill={rgb, 255:red, 184; green, 233; blue, 134 }  ,fill opacity=1 ] (152.33,31) -- (167.67,31) -- (167.67,91) -- (152.33,91) -- cycle ;
\draw  [fill={rgb, 255:red, 184; green, 233; blue, 134 }  ,fill opacity=1 ] (167.67,31) -- (183,31) -- (183,91) -- (167.67,91) -- cycle ;
\draw   (137,31) -- (183,31) -- (183,47) -- (137,47) -- cycle ;
\draw  [fill={rgb, 255:red, 248; green, 231; blue, 28 }  ,fill opacity=1 ] (220.45,111.31) -- (271.26,126) -- (271.85,184.8) -- (221.34,200.5) -- cycle ;
\draw  [fill={rgb, 255:red, 184; green, 233; blue, 134 }  ,fill opacity=1 ] (312,131) -- (327.33,131) -- (327.33,191) -- (312,191) -- cycle ;
\draw  [fill={rgb, 255:red, 80; green, 227; blue, 194 }  ,fill opacity=1 ] (345.33,132) -- (360.67,132) -- (360.67,192) -- (345.33,192) -- cycle ;
\draw  [fill={rgb, 255:red, 74; green, 144; blue, 226 }  ,fill opacity=1 ] (376,132) -- (391.33,132) -- (391.33,192) -- (376,192) -- cycle ;
\draw    (186,156) -- (207,156) ;
\draw [shift={(210,156)}, rotate = 180] [fill={rgb, 255:red, 0; green, 0; blue, 0 }  ][line width=0.08]  [draw opacity=0] (8.93,-4.29) -- (0,0) -- (8.93,4.29) -- cycle    ;
\draw    (279,155) -- (300,155) ;
\draw [shift={(303,155)}, rotate = 180] [fill={rgb, 255:red, 0; green, 0; blue, 0 }  ][line width=0.08]  [draw opacity=0] (8.93,-4.29) -- (0,0) -- (8.93,4.29) -- cycle    ;
\draw    (418,160) -- (495,160) ;
\draw [shift={(498,160)}, rotate = 180] [fill={rgb, 255:red, 0; green, 0; blue, 0 }  ][line width=0.08]  [draw opacity=0] (8.93,-4.29) -- (0,0) -- (8.93,4.29) -- cycle    ;
\draw    (188,99) -- (204.82,114.94) ;
\draw [shift={(207,117)}, rotate = 223.45] [fill={rgb, 255:red, 0; green, 0; blue, 0 }  ][line width=0.08]  [draw opacity=0] (8.93,-4.29) -- (0,0) -- (8.93,4.29) -- cycle    ;
\draw    (188,220) -- (204.76,205) ;
\draw [shift={(207,203)}, rotate = 138.18] [fill={rgb, 255:red, 0; green, 0; blue, 0 }  ][line width=0.08]  [draw opacity=0] (8.93,-4.29) -- (0,0) -- (8.93,4.29) -- cycle    ;
\draw  [fill={rgb, 255:red, 80; green, 227; blue, 194 }  ,fill opacity=1 ] (536.33,137) -- (551.67,137) -- (551.67,197) -- (536.33,197) -- cycle ;

\draw (150,8) node [anchor=north west][inner sep=0.75pt]   [align=left] {$\displaystyle \mathbf{x}^{( 1)}$};
\draw (149,105) node [anchor=north west][inner sep=0.75pt]   [align=left] {$\displaystyle \mathbf{x}^{( 2)}$};
\draw (149,202) node [anchor=north west][inner sep=0.75pt]   [align=left] {$\displaystyle \mathbf{x}^{( 3)}$};
\draw (309,109) node [anchor=north west][inner sep=0.75pt]   [align=left] {$\displaystyle \mathbf{z}^{( 1)}$};
\draw (340,110) node [anchor=north west][inner sep=0.75pt]   [align=left] {$\displaystyle \mathbf{z}^{( 2)}$};
\draw (372,111) node [anchor=north west][inner sep=0.75pt]   [align=left] {$\displaystyle \mathbf{z}^{( 3)}$};
\draw (239,149) node [anchor=north west][inner sep=0.75pt]  [font=\small] [align=left] {$\displaystyle \mathbf{E}$};
\draw (136,34) node [anchor=north west][inner sep=0.75pt]  [font=\scriptsize] [align=left] {$\displaystyle A$};
\draw (166.67,34) node [anchor=north west][inner sep=0.75pt]  [font=\scriptsize] [align=left] {$\displaystyle C$};
\draw (165.67,131) node [anchor=north west][inner sep=0.75pt]  [font=\scriptsize] [align=left] {$\displaystyle C$};
\draw (135,131) node [anchor=north west][inner sep=0.75pt]  [font=\scriptsize] [align=left] {$\displaystyle A$};
\draw (135,227) node [anchor=north west][inner sep=0.75pt]  [font=\scriptsize] [align=left] {$\displaystyle A$};
\draw (165.67,227) node [anchor=north west][inner sep=0.75pt]  [font=\scriptsize] [align=left] {$\displaystyle C$};
\draw (151.33,34) node [anchor=north west][inner sep=0.75pt]  [font=\scriptsize] [align=left] {$\displaystyle B$};
\draw (150.33,131) node [anchor=north west][inner sep=0.75pt]  [font=\scriptsize] [align=left] {$\displaystyle B$};
\draw (150.33,227) node [anchor=north west][inner sep=0.75pt]  [font=\scriptsize] [align=left] {$\displaystyle B$};
\draw (538,120) node [anchor=north west][inner sep=0.75pt]   [align=left] {$\displaystyle \mathbf{z}$};
\draw (410,134) node [anchor=north west][inner sep=0.75pt]   [align=left] {$\displaystyle \mathbf{Aggregation}$};

\end{tikzpicture}

    \caption{Illustration of aggregation of representations at test time}
    \label{fig:agg}
\end{figure}

Latent representations, $\mathbf{z}^1$, $\mathbf{z}^2, \cdots, \mathbf{z}^M$ are aggregated by mean (default), max, min or sum as shown in \ref{fig:agg}. The aggregated representation is then used for the classification task. The classification loss is defined in Equation \ref{eq:class}.

\subsection{Adding subset identity}
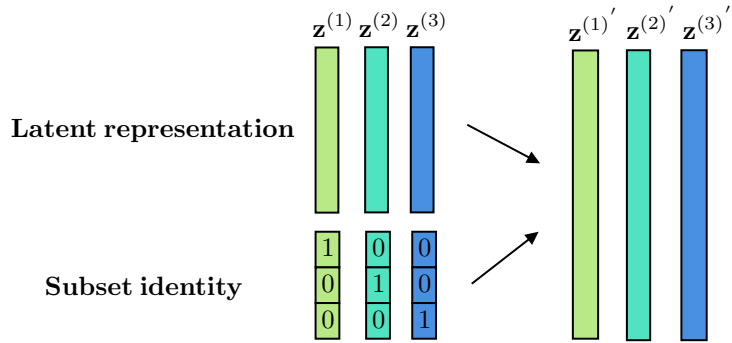
\begin{figure}[h]
    \centering
    \begin{tikzpicture}[x=0.75pt,y=0.75pt,yscale=-0.75,xscale=0.75]

\draw  [fill={rgb, 255:red, 184; green, 233; blue, 134 }  ,fill opacity=1 ] (257,63) -- (272.33,63) -- (272.33,174.15) -- (257,174.15) -- cycle ;
\draw  [fill={rgb, 255:red, 80; green, 227; blue, 194 }  ,fill opacity=1 ] (290.33,62.85) -- (305.67,62.85) -- (305.67,174) -- (290.33,174) -- cycle ;
\draw  [fill={rgb, 255:red, 74; green, 144; blue, 226 }  ,fill opacity=1 ] (321,62.85) -- (336.33,62.85) -- (336.33,174) -- (321,174) -- cycle ;
\draw  [fill={rgb, 255:red, 184; green, 233; blue, 134 }  ,fill opacity=1 ] (257,187) -- (273,187) -- (273,211) -- (257,211) -- cycle ;
\draw  [fill={rgb, 255:red, 184; green, 233; blue, 134 }  ,fill opacity=1 ] (257,211) -- (273,211) -- (273,235) -- (257,235) -- cycle ;
\draw  [fill={rgb, 255:red, 184; green, 233; blue, 134 }  ,fill opacity=1 ] (257,235) -- (273,235) -- (273,259) -- (257,259) -- cycle ;
\draw  [fill={rgb, 255:red, 80; green, 227; blue, 194 }  ,fill opacity=1 ] (291,187) -- (307,187) -- (307,211) -- (291,211) -- cycle ;
\draw  [fill={rgb, 255:red, 80; green, 227; blue, 194 }  ,fill opacity=1 ] (291,211) -- (307,211) -- (307,235) -- (291,235) -- cycle ;
\draw  [fill={rgb, 255:red, 80; green, 227; blue, 194 }  ,fill opacity=1 ] (291,235) -- (307,235) -- (307,259) -- (291,259) -- cycle ;
\draw  [fill={rgb, 255:red, 74; green, 144; blue, 226 }  ,fill opacity=1 ] (322,187) -- (338,187) -- (338,211) -- (322,211) -- cycle ;
\draw  [fill={rgb, 255:red, 74; green, 144; blue, 226 }  ,fill opacity=1 ] (322,211) -- (338,211) -- (338,235) -- (322,235) -- cycle ;
\draw  [fill={rgb, 255:red, 74; green, 144; blue, 226 }  ,fill opacity=1 ] (322,235) -- (338,235) -- (338,259) -- (322,259) -- cycle ;
\draw    (359,115) -- (405.35,139.59) ;
\draw [shift={(408,141)}, rotate = 207.95] [fill={rgb, 255:red, 0; green, 0; blue, 0 }  ][line width=0.08]  [draw opacity=0] (8.93,-4.29) -- (0,0) -- (8.93,4.29) -- cycle    ;
\draw    (362,222) -- (404.63,188.84) ;
\draw [shift={(407,187)}, rotate = 142.13] [fill={rgb, 255:red, 0; green, 0; blue, 0 }  ][line width=0.08]  [draw opacity=0] (8.93,-4.29) -- (0,0) -- (8.93,4.29) -- cycle    ;
\draw  [fill={rgb, 255:red, 184; green, 233; blue, 134 }  ,fill opacity=1 ] (430,65) -- (447,65) -- (447,258.99) -- (430,258.99) -- cycle ;
\draw  [fill={rgb, 255:red, 80; green, 227; blue, 194 }  ,fill opacity=1 ] (466.33,64.85) -- (482,64.85) -- (482,260) -- (466.33,260) -- cycle ;
\draw  [fill={rgb, 255:red, 74; green, 144; blue, 226 }  ,fill opacity=1 ] (503,64.85) -- (520,64.85) -- (520,260) -- (503,260) -- cycle ;

\draw (253,37) node [anchor=north west][inner sep=0.75pt]   [align=left] {$\displaystyle \mathbf{z}^{( 1)}$};
\draw (284,37) node [anchor=north west][inner sep=0.75pt]   [align=left] {$\displaystyle \mathbf{z}^{( 2)}$};
\draw (316,37) node [anchor=north west][inner sep=0.75pt]   [align=left] {$\displaystyle \mathbf{z}^{( 3)}$};
\draw (259,190) node [anchor=north west][inner sep=0.75pt]   [align=left] {$\displaystyle 1$};
\draw (259,214) node [anchor=north west][inner sep=0.75pt]   [align=left] {$\displaystyle 0$};
\draw (259,238) node [anchor=north west][inner sep=0.75pt]   [align=left] {$\displaystyle 0$};
\draw (293,190) node [anchor=north west][inner sep=0.75pt]   [align=left] {$\displaystyle 0$};
\draw (293,214) node [anchor=north west][inner sep=0.75pt]   [align=left] {$\displaystyle 1$};
\draw (293,238) node [anchor=north west][inner sep=0.75pt]   [align=left] {$\displaystyle 0$};
\draw (324,190) node [anchor=north west][inner sep=0.75pt]   [align=left] {$\displaystyle 0$};
\draw (324,214) node [anchor=north west][inner sep=0.75pt]   [align=left] {$\displaystyle 0$};
\draw (324,238) node [anchor=north west][inner sep=0.75pt]   [align=left] {$\displaystyle 1$};
\draw (425,34) node [anchor=north west][inner sep=0.75pt]   [align=left] {$\displaystyle \mathbf{z}{^{( 1)}}^{'}$};
\draw (73,215) node [anchor=north west][inner sep=0.75pt]   [align=left] {$\displaystyle \mathbf{Subset\ identity}$};
\draw (50,109) node [anchor=north west][inner sep=0.75pt]   [align=left] {$\displaystyle \mathbf{Latent\ representation}$};
\draw (464,33) node [anchor=north west][inner sep=0.75pt]   [align=left] {$\displaystyle \mathbf{z}{^{( 2)}}^{'}$};
\draw (502,33) node [anchor=north west][inner sep=0.75pt]   [align=left] {$\displaystyle \mathbf{z}{^{( 3)}}^{'}$};

\end{tikzpicture}

    \caption{Illustration of adding subset identity. Latent representations $\mathbf{z}^{(1)}$, $\mathbf{z}^{(2)}$ and $\mathbf{z}^{3)}$ are concatenated with their corresponding subset identity vectors to produce ${\mathbf{z}^{(1)}}^{'}$, ${\mathbf{z}^{(2)}}^{'}$ and ${\mathbf{z}^{(3)}}^{'}$. These vectors are then input to decoder and downstream layers.}
    \label{fig:subid}
\end{figure}

To help the model identify which subset is fed into it, we add a one-hot encoded subset identity. Once the latent representation $\mathbf{z}^j$ is produced by the encoder, it is concatenated with the one-hot encoded subset identity to produce ${\mathbf{z}^j}^{'}$ as shown in Figure \ref{fig:subid}. This concatenated vector is then passed onto the decoder and downstream layers. 

In the setting where subset identity is used, the aggregation of representation at test time is not required. Therefore, the downstream layer output from any one of the subsets is used as the model output. We noticed that aggregating the downstream outputs produced by representations from multiple subsets performed similar to the setting where representation form any one of the subsets is used to produce downstream output. This shows the power of adding subset identity.

\section{Experiments and results}
\subsection{Implementation details}
Our codebase (\href{https://github.com/hashimsayed0/OmiEmbed}{https://github.com/hashimsayed0/OmiEmbed}) is built upon the codebase of the baseline that we implemented \cite{omiembed} (\href{https://github.com/zhangxiaoyu11/OmiEmbed}{https://github.com/zhangxiaoyu11/OmiEmbed}). The framework is built on PyTorch \cite{pytorch}. It supports multiple omics modes, backbone networks, reconstruction losses and latent space dimensions. The model was tuned on each of these hyperparameters. The following sections describe the experiments in detail.

\subsection{Baseline experiments}
\subsubsection{Handling missing values}
The DNA methylation dataset had missing values. We tried various approaches to handle them and evaluated the model's performance on cancer type classification. Zero \& mean imputations as well as dropping features containing missing values were the three approaches attempted. Mean imputation provided best performance as shown in Table \ref{tab:imp}. Therefore, mean imputed DNA methylation dataset was used for further experiments.
\begin{table}[h]
\begin{tabular}{lccccc}
\hline
\textbf{Imputation method} & \textbf{Accuracy} & \textbf{Precision} & \textbf{Recall} & \textbf{F1}     & \textbf{AUC}    \\ \hline
Dropping features          & 0.9512            & 0.9387             & 0.9065          & 0.9190          & 0.9976          \\
Zero imputation            & 0.9517            & 0.9421             & 0.9256          & 0.9305          & \textbf{0.9987} \\
\textbf{Mean imputation}   & \textbf{0.9589}   & \textbf{0.9470}    & \textbf{0.9552} & \textbf{0.9503} & 0.9981          \\ \hline
\end{tabular}
\centering
\caption{Effect of various imputation methods on cancer type classification performance with only DNA methylation dataset}
\label{tab:imp}
\end{table}

\subsubsection{Omics mode}
\begin{figure}[h]
    \centering
    \includegraphics[scale=0.4]{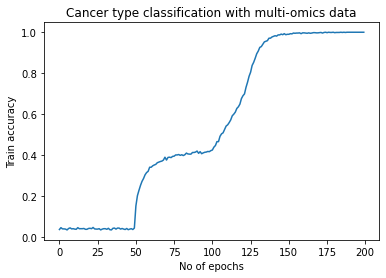}
    \caption{Downstream task train accuracy with multi-omics data (A+B+C) in 3 phases. Phase 1 and 2 are 50 epochs while phase 3 is 100 epochs.}
    \label{fig:train}
\end{figure}

For this set of experiments, we tried multiple combinations of omics data types. As shown in Table \ref{tab:mode}, using all three data types showed best performance. This can be because each type of data reveals some kind of information about the tumour, which eventually helps the model classify tumours. Also, note that miRNA expression dataset with under 2000 features alone gives almost 95\% accuracy in the task, showing the predictive capacity of omics data. Figure \ref{fig:train} shows how training evolves through the 3 phases discussed earlier.

\begin{table}[h]
\begin{tabular}{lccccc}
\hline
\textbf{Omics mode}          & \textbf{Accuracy} & \textbf{Precision} & \textbf{Recall} & \textbf{F1}     & \textbf{AUC}    \\ \hline
RNA-seq gene expression (A)          & 0.9584            & 0.9532             & 0.9283          & 0.9257          & 0.9986          \\
DNA methylation data (B)          & 0.9589            & 0.9470             & 0.9552          & 0.9503          & 0.9981          \\
miRNA stem loop expression (C)         & 0.9494            & 0.9355             & 0.9254          & 0.9286          & 0.9984          \\
Multi omics (A+B)            & 0.9647            & 0.9506             & 0.9497          & 0.9497          & 0.9973          \\
\textbf{Multi omics (A+B+C)} & \textbf{0.9709}   & \textbf{0.9640}    & \textbf{0.9597} & \textbf{0.9613} & \textbf{0.9988} \\ \hline
\end{tabular}
\centering
\caption{Evaluation of OmiEmbed on cancer type classification with multiple omics type combinations}
\label{tab:mode}
\end{table}

\subsubsection{Backbone of the network}
Two main backbones that we trained for the encoder decoder architecture are fully connected and convolutional networks. In the fully connected network, DNA methylation data was divided into 23 blocks, corresponding to the chromosome number of respective CpG sites. This reduces the number of parameters, avoids overfitting and uses less GPU memory, while also making the model learn inter-chromosome relationships in first hidden layer and intra-chromosome relationships in subsequent layers. As shown in Table \ref{tab:back}, the extra chromosome separation step improves performance.

\begin{table}[h]
\begin{tabular}{lccccc}
\hline
\textbf{Backbone}                   & \textbf{Accuracy} & \textbf{Precision} & \textbf{Recall} & \textbf{F1}     & \textbf{AUC}    \\ \hline
Conv-1D                             & 0.9536            & 0.9345             & 0.9239          & 0.9278          & 0.9971          \\
Fully connected (FC)                & 0.9598            & 0.9302             & 0.9388          & 0.9336          & 0.9983          \\
\textbf{FC + chromosome separation} & \textbf{0.9626}   & \textbf{0.9561}    & \textbf{0.9488} & \textbf{0.9443} & \textbf{0.9985} \\ \hline
\end{tabular}
\centering
\caption{Evaluation of OmiEmbed on cancer type classification with various backbones for embedding module}
\label{tab:back}
\end{table}

\subsubsection{Latent space dimension}
\begin{figure}[h]
    \centering
    \includegraphics[scale=0.6]{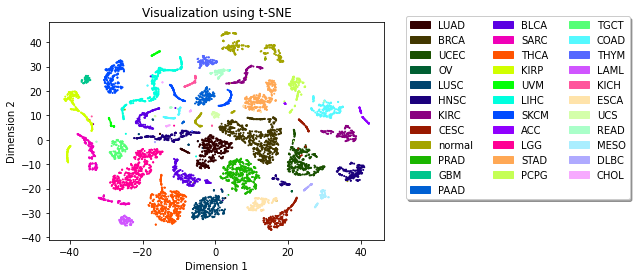}
    \caption{Visualizing the 128D latent vector learned by the end of phase 3 training using t-SNE. It can be seen that the classes are separated. Labels on the right side refer to the TCGA short codes for cancer types}
    \label{fig:tsne}
\end{figure}

This set of experiments was aimed at choosing the ideal dimension of latent space. We experimented with 64, 128 and 256. As shown in Table \ref{tab:lat}, 128 dimensional latent space gives best performance. This could be due to the fact that its large enough to embed all the essential information. The 128D latent vector was visualized using t-SNE \cite{tsne} in Figure \ref{fig:tsne}, which shows that the network learns to discriminate between various tumour types and the normal class. Even classes with few number of samples are separated by the model.

\begin{table}[h]
\begin{tabular}{lccccc}
\hline
\textbf{Latent space dimension} & \textbf{Accuracy} & \textbf{Precision} & \textbf{Recall} & \textbf{F1}     & \textbf{AUC}    \\ \hline
64                              & 0.9693            & 0.9588             & 0.9578          & 0.9568          & 0.9981          \\
256                             & 0.9631            & 0.9398             & \textbf{0.9618} & 0.9473          & 0.9983          \\
\textbf{128}                    & \textbf{0.9709}   & \textbf{0.9640}    & 0.9597          & \textbf{0.9613} & \textbf{0.9988} \\ \hline
\end{tabular}
\centering
\caption{Evaluation of OmiEmbed on cancer type classification with various latent space dimensions}
\label{tab:lat}
\vspace{-4mm}
\end{table}

\subsubsection{Reconstruction loss}
We also tried different reconstruction losses. The results are shown in Table \ref{tab:rec}. Note that binary cross entropy loss outperforms means squared error loss in all metrics except accuracy. 

\begin{table}[h]
\begin{tabular}{lccccc}
\hline
\textbf{Reconstruction loss}  & \textbf{Accuracy} & \textbf{Precision} & \textbf{Recall} & \textbf{F1}     & \textbf{AUC}    \\ \hline
Mean absolute error (L1)      & 0.9620            & 0.9542             & 0.9430          & 0.9407          & 0.9977          \\
Mean squared error            & \textbf{0.9648}   & 0.9506             & 0.9374          & 0.9399          & 0.9979          \\
\textbf{Binary cross entropy} & 0.9626            & \textbf{0.9561}    & \textbf{0.9488} & \textbf{0.9443} & \textbf{0.9985} \\ \hline
\end{tabular}
\centering
\caption{Evaluation of OmiEmbed on cancer type classification with various reconstruction losses}
\label{tab:rec}
\end{table}

\subsection{Feature subsetting experiments}
\subsubsection{Reducing the network size}
Once we applied the features subsetting approach, we noticed that the network was overfitting to the subsets. This is because we were using the same network as used in the previous experiments with just the change in input dimension. To solve this issue, we reduced the dimensions of encoder, decoder and downstream layers by encoder reduction factor, decoder reduction factor and downstream reduction factors respectively compared to the baseline network. This set of experiments were run to choose the optimal network size. Results are shown in Table \ref{tab:red}. It is striking to see that with feature subsetting, a network with reduced size achieves comparable results to the baseline. Due to the success of networks with lower of sizes, the encoder, decoder and downstream layers of the network were reduced by 8, 4 and 2 in the following experiments. 

\begin{table}[h]
\begin{tabular}{lcccccc}
\hline
\textbf{Reduction} & \textbf{Accuracy} & \textbf{Precision} & \textbf{Recall} & \textbf{F1}    & \textbf{ROC AUC} & \textbf{Mean metric} \\ \hline
1\_1\_1            & 0.89              & 0.881              & 0.908           & 0.859          & 0.997            & 0.893                \\
1\_1\_2            & 0.922             & 0.925              & 0.894           & 0.879          & 0.998            & 0.914                \\
1\_4\_1            & 0.931             & 0.902              & 0.941           & 0.905          & 0.999            & 0.925                \\
1\_4\_2            & 0.91              & 0.892              & 0.92            & 0.888          & 0.998            & 0.907                \\
4\_1\_1            & \textbf{0.962}    & 0.93               & 0.909           & 0.909          & 0.999            & 0.934                \\
4\_1\_2            & 0.949             & 0.933              & 0.911           & 0.916          & 0.994            & 0.931                \\
4\_4\_1            & 0.954             & 0.937              & 0.939           & 0.929          & 0.999            & 0.943                \\
4\_4\_2            & 0.943             & 0.893              & 0.91            & 0.898          & 0.997            & 0.915                \\
8\_1\_1            & 0.945             & 0.927              & 0.935           & 0.928          & 0.999            & 0.936                \\
8\_1\_2            & 0.954             & 0.945              & \textbf{0.943}  & 0.94           & \textbf{0.999}   & \textbf{0.947}       \\
8\_4\_1            & 0.944             & \textbf{0.947}     & 0.907           & 0.907          & 0.999            & 0.933                \\
8\_4\_2            & 0.945             & 0.942              & 0.935           & \textbf{0.936} & 0.999            & 0.941                \\ \hline
\end{tabular}\\
\centering
\caption{Evaluation of SubOmiEmbed on cancer type classification with multiple reduction factors. The dimension reduction factors are mentioned in the first column. For instance, 8\_4\_2 implies that the dimensions of encoder, decoder and downstream layers were reduced by 8, 4 and 2 respectively compared to the baseline network. Please note that the mean metric mentioned in this table as well as subsequent tables refer to the mean of accuracy, precision and recall.}
\label{tab:red}

\end{table}

\subsubsection{Aggregation method}
In this set of experiments, we aggregated the representations at test time with different methods. Adding subset identity didn't require aggregation and it was also compared against the other methods. A random subset with identity concatenated with it was used as input to decoder and downstream layers in this setting. The results are shown in Table \ref{tab:agg}. From the results, it can be seen that summing up the subset is the best aggregation method to use. Perhaps this way, not much information is lost during aggregation. Another insight is that choosing a random subset with one-hot encoded identity vector performs much better than without it. Due to the efficiency of using just one random subset with identity, we use it in our following experiments.

\begin{table}[h]
\begin{tabular}{lcccccc}
\hline
\textbf{Aggregation method} & \textbf{Accuracy} & \textbf{Precision} & \textbf{Recall} & \textbf{F1}    & \textbf{ROC AUC} & \textbf{Mean metric} \\ \hline
Maximum                     & 0.883             & 0.88               & 0.905           & 0.873          & 0.998            & 0.889                \\
Mean                        & 0.936             & 0.939              & 0.933           & 0.93           & 0.999            & 0.936                \\
Minimum                     & 0.935             & 0.907              & 0.932           & 0.913          & 0.997            & 0.925                \\
Random subset               & 0.897             & 0.869              & 0.814           & 0.825          & 0.996            & 0.86                 \\
Random with identity        & 0.945             & 0.942              & 0.935           & 0.936          & 0.999            & 0.941                \\
\textbf{Sum}                & \textbf{0.948}    & \textbf{0.951}     & \textbf{0.942}  & \textbf{0.942} & \textbf{0.999}   & \textbf{0.947}       \\ \hline
\end{tabular}\\
\centering
\caption{Evaluation of SubOmiEmbed on cancer type classification with multiple aggregation methods. Using a random subset with subset identity was also tested in this set of experiments.}
\label{tab:agg}
\end{table}

\subsubsection{Number of subsets}
Another set of experiments we conducted was to choose the optimal number of subsets. Results are shown in Table \ref{tab:subnum}. It is surprising to see that the network with reduced dimensions works well with even 10 subsets. This shows the power of feature subsetting.

\begin{table}[h]
\begin{tabular}{lcccccc}
\hline
\textbf{No of subsets} & \textbf{Accuracy} & \textbf{Precision} & \textbf{Recall} & \textbf{F1}    & \textbf{ROC AUC} & \textbf{Mean metric} \\ \hline
2                      & 0.959             & \textbf{0.96}      & 0.94            & 0.944          & 0.999            & 0.953                \\
3                      & 0.957             & 0.946              & 0.939           & 0.927          & 0.999            & 0.947                \\
4                      & \textbf{0.963}    & 0.95               & \textbf{0.958}  & \textbf{0.952} & \textbf{0.999}   & \textbf{0.957}       \\
5                      & 0.959             & 0.913              & 0.909           & 0.903          & 0.999            & 0.927                \\
6                      & 0.951             & 0.946              & 0.953           & 0.942          & 0.999            & 0.95                 \\
7                      & 0.943             & 0.915              & 0.921           & 0.891          & 0.999            & 0.926                \\
8                      & 0.93              & 0.934              & 0.933           & 0.922          & 0.999            & 0.932                \\
9                      & 0.948             & 0.955              & 0.926           & 0.936          & 0.998            & 0.943                \\
10                     & 0.95              & 0.941              & 0.944           & 0.94           & 0.999            & 0.945                \\
11                     & 0.947             & 0.931              & 0.932           & 0.924          & 0.999            & 0.937                \\
12                     & 0.939             & 0.952              & 0.92            & 0.927          & 0.999            & 0.937                \\ \hline
\end{tabular}\\
\centering
\caption{Evaluation of SubOmiEmbed on cancer type classification with various number of subsets. Note that all experiments were run on a network with reduced size and with subset identity.}
\label{tab:subnum}
\end{table}
\subsection{Comparison to other methods}
The performance of OmiEmbed on cancer type classification compared to other methods are shown in Table \ref{tab:comp}. As can be seen, OmiEmbed outperforms other methods on this task. 

\begin{table}[h]
\begin{tabular}{lccccc}
\hline
\textbf{Method} & \textbf{Accuracy} & \textbf{Precision} & \textbf{Recall} & \textbf{F1}     & \textbf{ROC AUC} \\ \hline
kNN             & 0.898             & 0.89               & 0.87            & 0.87            & -                \\
SVM             & 0.912             & 0.88               & 0.87            & 0.87            & -                \\
NN              & 0.935             & 0.937              & 0.932           & -               & 0.986            \\
GCN \cite{ricardo}            & 0.946             & 0.947              & 0.921           & -               & -                \\
CNN \cite{Lyu}            & 0.955             & \textbf{0.955}     & \textbf{0.955}  & 0.954           &                  \\
OmiEmbed       & \textbf{0.977}   & -                  & -               & \textbf{0.968} & \textbf{0.999}  \\ \hline
\end{tabular} \\
\centering 
\caption{Evaluation of various methods on cancer type classification. NN refers to a simple neural network with few hidden layers. GCN \cite{ricardo} refers to a work that involves computing gene co-expression graph and training graph convolutional network on it. CNN \cite{Lyu} refers to a paper that embeds the data into 2-D images and uses a convolutional neural network for classification.}
\label{tab:comp}
\end{table}

\section{Conclusions}
In this project, we have explored the area of representation learning for multi-omics data. We began with the study of the biological relevance of the dataset, and looked for deep learning models to extract meaningful representations from the different types of omics profiling. We then identified the field of VAEs and its applications. After formulating the problem and loss functions, we ran extensive experiments to evaluate the learning capacity of VAEs. In our experiments, we pre-trained the embedding module on reconstruction of training data in the first phase, then kept the embedding layers frozen and trained it for the downstream task of tumour type classification in the second phase. In the last phase, we trained the model on both tasks concurrently to produce meaningful representation of the input data that is relevant for the downstream task. Various insights about the model and its hyperparameters were derived from the experiments.

After experimenting with the baseline model, we extended our framework to include the SSL technique of feature subsetting as well as adding subset identity. We first formulated the problem and then ran our experiments. There were challenges along the way, one of which being overfitting of the model to subsets. We reduced the network size to reduce overfitting in addition to giving the model information about the ordering of the subsets. With our experiments, we show that a model of a much reduced size can produce comparable results to that of the baseline model, using a single subset of the data. This shows that model learns a lot more with self supervision. This work can be extended to add noise to the data, and amending the embedding loss to include contrastive loss. Attention based techniques could also be used to integrate sequence based genomic data into the framework.


\clearpage
\bibliographystyle{abbrv}
\bibliography{ref.bib}
\end{document}